# SWift - A SignWriting improved fast transcriber[1]

### C.S. Bianchini
Université Paris 8/UMR7023 - 2 rue de la Liberté, Saint-Denis, Paris (France) - chiadu14@tiscali.it

### F. Borgia
Université Toulouse III - 118 Route de Narbonne, Toulouse (France) - fabrizio.borgia@uniroma1.it

### P. Bottoni, M. De Marsico
Sapienza Università di Roma, Department of Computer Science - Via Salaria 113, Rome - fbottoni/demarsicog@di.uniroma1.it

**ABSTRACT**
We present SWift (SignWriting improved fast transcriber), an advanced editor for computer-aided writing and transcribing using SignWriting (SW). SW is devised to allow deaf people and linguists alike to exploit an easy-to-grasp written form of (any) sign language. Similarly, SWift has been developed for everyone who masters SW, and is not exclusively deaf-oriented. Using SWift, it is possible to compose and save any sign, using elementary components called glyphs. A guided procedure facilitates the composition process. SWift is aimed at helping to break down the "electronic" barriers that keep the deaf community away from Information and Communication Technology (ICT). The editor has been developed modularly and can be integrated everywhere the use of SW, as an alternative to written vocal language, may be advisable.

**Categories and Subject Descriptors**
H.5.2 [User Interfaces]: User-centered design, accessibility, users with special needs
H.5.2 [User Interfaces]: Interaction styles
H.5.2 [User Interfaces]: Input devices and strategies

**General Terms**
Human factors

## 1. INTRODUCTION
Deaf people are 0.1% of the world population, and most of them, even if they know a vocal language (VL), communicate through a visual-gestural form of their national Sign Language (SL). The structure of SL is deeply different from the sequential structure of VL due to an intensive use of spatial-temporal (multi-linear) relationships. Recent studies stressed the importance of the manual aspects of signs as well as the importance of information conveyed by other components [1 Cuxac and Antinoro Pizzuto, 2010], such as gaze, facial expression and body movements. Research on SL led to question the representativeness of the language, as they do not have historically defined

---

writing systems. Most representations of SL are inadequate to effectively encode the non-manual components, limiting their ability to show all the elements constituting and signifying SL [2 Fajardo et al., 2009]. Besides this discussion on the scientific representation of SL, the need to write down (and not only to transcribe, as it is the case in research) these languages must also be taken into account. Recent studies [3 Garcia and Perini, 2010] have demonstrated the difficulties encountered by deaf people in learning and using written VL, which severely limits the access to written texts as well as their production. Therefore, a "written" form closer to SL as deafs' most natural way of expression would be of great importance. The deaf community has recently replaced many functions of \writing" by videos, but this is not always possible; for example, it is not possible to take notes with a video, or to formulate a query on a search engine. So far, the only noteworthy attempts to address accessibility issues pertain to the Web Content Accessibility Guidelines (WCAG) document - http://www.w3.org - a set of guidelines directed to digital developers to make their resources accessible to people with disabilities. However, there is poor knowledge of problems encountered by deaf people in "decoding" written VL [4 Perfetti and Sandak, 2000]: subtitles are considered a sufficient support, causing deaf people to face severe accessibility barriers. WCAG2.0 (2008) better addresses the above problems, trying to increase the perceivability ofWeb content - even by deaf people - through instructions and technical suggestions. These criteria still focus on the transcription of audio content, though. The integration of information with SL videos [3 Garcia and Perini, 2010] is usually part of approaches to create whole SL-tongued web sites. Such solutions present two main drawbacks. First, it is often unfeasible to set up recording of a high number of clips to transmit the complete information. Second, video fruition on the web if often hindered by bandwidth limitations. A written form for SL would solve many of the above problems; a promising candidate is SignWriting (SW). The goal of our work is to make it effectively exploitable as a communication mean, and as a suitable learning support for deafs. To this aim, we developed SWift with the advice of experts and deaf researchers from ISTC-CNR (Institute of Cognitive Sciences and Technologies of the Italian National Research Council). This Web application currently allows users to compose single signs, in a simple interface-supported way.

## 2. ABOUT SIGNWRITING

SignWriting (SW) (http://www.signwriting.org) [5 Sutton, 1995] is a graphical framework, which allows reproduction of isolated signs as well as of discourse by composing basic elements (called glyphs), representing both manual and non-manual components of signs. Its strength is the ability to frame a rich coding system within a bi-dimensional representation, relying on iconic symbols. The set of glyphs (the International SW Alphabet - ISWA) may be considered as an alphabet that can be used to write any national SL. ISWA is composed by about 35,000 images (.png), each representing a single glyph, which is in turn identified by a unique code describing its features (see Fig. 1).

Figure 1: SW for LIS sign "various": labels are top-bottom and contact glyph is between hands.

Despite the apparent complexity of the notation, SW is very easy to learn for a deaf user, since it closely visually resembles the concrete signs. Moreover a SW transcription of text in any VL may also help with learning the VL itself. Bianchini [6 Bianchini, 2012] re-classified ISWA to improve consistency and effectiveness. She divided ISWA in categories, families and sub-families, which contain prototypical glyphs that are "declined" according to rules. **Categories** divide ISWA according to anatomical areas and kinds of elements to code. They are further divided into **Families** which are in turn divided into **Subfamilies**. Each glyph complies with the rules for its family and subfamily, with no possible exception. Categories, families and subfamilies are almost always derived from a different division of Sutton's groups and categories. Not all details of this new classification have been used for SWift, to avoid an excessive number of choices for the user and to keep the interface easy to understand and use.

## 3. ANALISYS OF SIGN MAKER (SM)

SWift was designed to allow digital content editing using SW. Of course, we first investigated related research projects. Valerie Sutton's own team developed SignMaker (SM) (see Fig. 2), a web application allowing writing and saving signs. The **Sign Display** is a whiteboard which shows the sign the user is composing. The **Glyph Menu** allows choosing a glyph, after searching it through a tree of submenus. The **Toolbox** contains buttons to perform tasks such as editing a single selected glyph, or resetting the Sign Display. Glyphs are dragged one at a time on the whiteboard from the Glyph Menu. They remain draggable, and become selectable one at a time, to be handled using Toolbox functions. The choice of freely draggable glyphs (*versus* glyphs in fixed positions) is consistent with SW lack of constraints, along the line of *natural interaction*, allowing users to perform actions as they would do without a computer. However, it is possible to select only one glyph at a time, so that simultaneous multiple editing is impossible. Moreover, the functions for managing the Sign Display are located in the Toolbox, hidden among other functions and far from their target.

Figure 2: Home screen of SM (highlighted areas).

The Glyph Menu allows browsing the ISWA through a tree: the user navigates from the "root" to a "leaf" menu, by choosing the desired types and subtypes of glyph. Each "leaf" contains a collection of glyphs (in the order of tens) from which the user can pick the desired one (Fig. 3). During a SM session, the user spends most of the time navigating the Glyph Menu, due to its structure and to the kind of interaction it provides (linear navigation along levels).

Figure 3: Tree-like structure of SM Glyph Menu.

The most critical problem is the spatial layout of the navigation menu. Glyph types are presented next to each other, without a clear distinction, and no graphic cue guides a user with a medium-poor knowledge of SW in his search path. The Toolbox provides a number of functions for handling any interactive component of SM interface. We observed SM major issues in this area, relating to the arrangement of the buttons and their content. Navigation buttons in the Glyph Menu are placed aside and mingle with those for glyph handling (the majority), and with those for handling the Sign Display itself. The new user barely orientates in this monolithic block, where the single buttons are stripped away from the target areas to which their functions apply. In addition to these issues, it is to consider that a consolidate guideline in human-computer interaction theory is to create self-explanatory icons, without the need for text. In our case this rule cannot be overlooked, having to do with users who may have difficulties in understanding text. Moreover, the text on the buttons is in English, and there is no way of choosing another language. In general, the expressiveness of the icons is acceptable, but all of them can be fully understood only by the small group of signing people with good knowledge of SW.

## 4. ANALISYS OF SWIFT

Fig. 4 shows the home screen of SWift. Such interface appears more friendly than that of SM: it minimizes the use of text labels and presents a collection of colorful and familiar icons.

Figure 4: Home screen of SWift (highlighted areas).

The goal is to make users feel comfortable and to avoid immediately confusing them with too much information, allowing immediate use of the program to a person accostumed with SW. Hence, the application was designed according to the core principles of User-Centered [7 Norman and Draper, 1986] and Contextual [8 Wixon et al., 1990] design. In fact, the Team of ISTC-CNR includes deaf researchers, who are a true sample of the main target users. Most choices derive from precise needs expressed by them. The redesign of the graphic interface has been supported by a complete rethinking of the application logic, so that we can actually consider SWift as a brand-new tool. The interface addresses some basic requirements:

- **Intuitiveness**: each function is presented in an intuitive and familiar way, since the user should not "learn"but rather "guess" how to use the interface.
- **Minimization of information**: each screen presents essential information, avoiding confusion.
- **Evocative icons**: the icons are familiar and large, without text; a user not understanding their meaning can however start an animation via mouse-over.
- **Minimization of text labels**: deaf people may have difficulties in correctly interpreting the meaning of text labels so we tried to keep text elements at a minimum [4 Perfetti and Sandak, 2000].
- **User-driven interface testing**: each major change of the interface has been discussed with the team of ISTC-CNR, and double-checked with the team itself, until the presentation of SWift was considered fully satisfying.

SWift will be analyzed like SM: the interface areas of interest will be considered separately.

The **Sign Display** has the same role as in SM. A first difference is that its handling is provided by two buttons on its immediate left. They allow to clear the Sign Display or to save the sign that has been composed in several formats: text, image, and SWift database. The latter includes the list of used glyphs, and allows statistical computation of compatibility, possibly providing hints to the user. More computations may allow researchers to further explore patterns in sign composition. Glyph searching through the **Glyph Menu** is based on a new intuitive system. We will discuss its details in the following. The **Toolbox** below the Sign Display provides several functions (rotation, duplication, erasing) to edit the glyph(s) that are currently selected in the Sign Display. Thanks to the multi-select feature provided by the Sign Display, more glyphs can be edited at once. The **Hint Panel** is one of themain innovations of SWift. Its task is to show the user a series of glyphs that are compatible with those in the Sign Display, based on the frequency of co-occurrences in the sign database. Statistics are updated immediately after any sign saved in SWift format. For sake of space, the Hint Panel is usually minimized, and is implemented as an expandable footer. When minimized, a textual label shows the number of compatible glyphs found. When maximized on user demand, it expands in a panel providing the ability to directly drag in the Sign Display glyphs proposed by the system, which exclusively belong to the anatomical area the user is currently searching.

Most design and optimization were focused on the Glyph Menu. Making the interaction with this menu more efficient reduces the search time of each glyph, which in turn determines composition time of each sign. The stylized human figure, named "Puppet" and the buttons beneath, present in a natural way the first important choice. The areas corresponding to any relevant anatomical element (head, shoulders, hands, arms) will highlight on mouse-over, suggesting the user the possibility to click and choose that area. By choosing one anatomic area of the Puppet, or one of the buttons which represent items such as punctuation and contacts, the user can access the dedicated search menu. As an example, clicking on the hands of the Puppet, the user will access the search menu for the hands, enabling the choice of configurations or motions related to this area. After the user clicks, the Puppet and the buttons beneath are reduced and shifted to the left, to form a navigation menu together with the button to return to the Glyph Menu home screen. To support user's orientation, a red square appears around a selected area, while the Puppet and the buttons remain clickable (Fig. 5). Hence, users can navigate from one area to another without returning to the home screen, which remains however reachable *via* an *ad hoc* button. In the central part of the menu, a label and an icon explain to the user what kind of glyphs are available in the group of boxes beneath. Each such blue rectangle, a "Choose Box", provides possible (incompatible) choices for a certain glyph feature. The user can look for the desired glyph according to any number of combined criteria. No predefined order constrains the sequence of choices, so that no line of reasoning is privileged. For example, a user might choose the glyph rotation first, while another could choose the number of fingers. Glyphs corresponding to the current selected criteria are placed into a scrollable panel and are immediately draggable into the Sign Display. Once an additional choice is made (or undone), the system selects the correct set of glyphs according to the updated feature values, and then the panel is updated accordingly. As an example, Fig. 5 shows the panel after choosing hand configurations with only one finger. In some cases the user can find the desired glyph by making only one choice; usually, about 3 features must be defined to find the searched glyph. In the worst case (i.e. selection of a checkbox in each Box) the user will have to choose between about 30 glyphs, still a number processable by a human in an acceptable time.

Figure 5: Glyph Menu - Choice made for hands.

## 5. USABILITY EVALUATION

We decided to test SWift adapting the Think Aloud Protocol. Of course, deaf users cannot "think aloud". They can rather express their thoughts through their SL, and often demonstrate a higher variability in their expressions. Roberts and Fels [9 Roberts and Fels, 2006] suggest a spatial setting where one camera records a rear view of the participant, the computer screen, and the interpreter while another camera records the front view of the participant, and the investigator. The problem with two cameras is to analyze and synchronize data from two recordings, with deriving difficulties to maintain a synoptic view of everything that happens at any given time. We therefore adopted a single camera, introducing the use of a projector. In this adaptation, the computer screen is projected on the wall, and the camera records anything worth of attention. Its oblique position with respect to the wall of the room minimizes the occlusion due to the monitor. The test was structured in three phases.

During the **welcome time**, the deaf participant (DP) is greeted and briefed by a screen containing a signed video and its transcription. The **sign-aloud test** is the most important phase; DPs are asked to perform a list of tasks to test SWift functions and usability, and to record anything that comes to her/his mind. The design of this phase deserved special attention. Users need reminders for which task is being performed, what needs to be done, etc. The task list should be available in both VL and SL. We chose to involve a hearing interpreter (HI), since the possibility of interaction between DPs and HI increases correct understanding of the tasks. In particular, HI always provided a task translation in SL at the beginning of each task. The tasks included both basic actions, such as inserting a random glyph, or looking for a particular one, and complex ones, such as composing a given sign. In the third phase, DPs were asked to answer to a usability questionnaire both in SL and in VL. The **final usability questionnaire** was designed by adapting the QUIS questionnaire to the needs of deaf users. In particular, each (simplified) written question was accompanied by the corresponding SL clip.

As for the first evaluation, ten deaf users were involved. Let us notice that it is generally difficult to perform a usability evaluation with a high number of users, when the target users' features are so specific (in our case, deaf and able to use SignWriting). Nevertheless, we are planning a wider experiment using the web. This requires further design efforts, in order to effectively exploit the remote test modality. The low number of errors with interface buttons made by DPs confirmed the quality of most of our choices. Some doubts regard the wastebasket icon, since it was seldom used. As for the navigation, the lack of any glyph in the home screen was misleading for most DPs, who expected to find them. In our opinion, this may be due to their former confidence with the SM interface. When facing a specific set of Choose Boxes, many DPs preferred to make a single choice, and then browse the whole resulting set (about 400 elements). We are working on better signaling the possibility of choosing multiple options. At the end of the test session, most DPs expressed appreciation for the test modalities, in particular for the final questionnaire. Its overall results were very satisfactory, and underlined precise trends for specific aspects of the application, so confirming the reliability of the obtained responses.

## 6. CONCLUSIONS

We have presented SWift, an advanced editor for SW. SWift allows users familiar with a Sign Language to compose and save a sign using elementary components (glyphs), through a friendly kind of interaction. The editor has been developed in a modular way, so it can be integrated everywhere the use of SW may be advisable, as an alternative to written VL. This might be help break down the "electronic" barriers that keep most deaf community away from ICT. The performed usability study achieved good results, and suggested further improvements for both the interface and interaction.

## 7. ACKNOWLEDGMENTS

This work was supported by Italian MIUR under the FIRB project "VISEL - E-learning, deafness, written language: a bridge of letters and signs towards knowledge society". We thank the deaf and hearing researchers at IST-CNR of Rome for their support and suggestions during the design and evaluation of SWift. In memory of Elena Antinoro Pizzuto.